\documentclass{article}

\usepackage{PRIMEarxiv}

\usepackage[utf8]{inputenc} 
\usepackage[T1]{fontenc}    
\usepackage{hyperref}       
\usepackage{url}            
\usepackage{booktabs}       
\usepackage{amsfonts}       
\usepackage{nicefrac}       
\usepackage{microtype}      
\usepackage{lipsum}
\usepackage{caption}
\usepackage{amsmath}
\usepackage{multirow}
\usepackage[numbers]{natbib}
 \usepackage{graphicx}
\usepackage{amssymb}
\usepackage{fancyhdr}       
\usepackage{graphicx}       
\graphicspath{{media/}}     
\newcommand\xlnet{\texttt{XLNet}}
\newcommand\transformerenc{\texttt{Transformer-Encoder}}
\newcommand\mask{\texttt{[MASK]}}
\newcommand\sep{\texttt{[SEP]}}
\title{Towards Answering Health-related Questions from Medical Videos: Datasets and Approaches
 }

\author{
  Deepak Gupta$^\dagger$, Kush Attal$^\ddagger$\thanks{Work done during Post-baccalaureate Fellow at NLM, NIH.}, and  Dina Demner-Fushman$^\dagger$ \\
  \newline
  $^\dagger$Lister Hill National Center for Biomedical Communications \\
  National Library of Medicine, National Institutes of Health \\
  Bethesda, MD, USA\\
  \texttt{\{firstname.lastname\}@nih.gov} \\
  \newline
 $^\ddagger$NYU Grossman School of Medicine \\
  New York University \\
  New York, NY, USA\\
  \texttt{Kush.Attal@nyulangone.org} 
}

\begin{document}
\maketitle

\begin{abstract}
The increase in the availability of online videos has transformed the way we access information and knowledge. A growing number of individuals now prefer instructional videos as they offer a series of step-by-step procedures to accomplish particular tasks. The instructional videos from the medical domain may provide the best possible visual answers to first aid, medical emergency, and medical education questions. Toward this, this paper is focused on answering health-related questions asked by the public by providing visual answers from medical videos. The scarcity of large-scale datasets in the medical domain is a key challenge that hinders the development of applications that can help the public with their health-related questions. To address this issue, we first proposed a pipelined approach to create two large-scale datasets: HealthVidQA-CRF and HealthVidQA-Prompt. Later, we proposed monomodal and multimodal approaches that can effectively provide visual answers from medical videos to natural language questions. We conducted a comprehensive analysis of the results, focusing on the impact of the created datasets on model training and the significance of visual features in enhancing the performance of the monomodal and multi-modal approaches. Our findings suggest that these datasets have the potential to enhance the performance of medical visual answer localization tasks and provide a promising future direction to further enhance the performance by using pre-trained language-vision models.
\end{abstract}

\section{Introduction} \label{sec:intro}
An effective multimodal system that can enhance the ability to interact with the visual world, which encompasses images and videos, using a natural language query, has always been a coveted goal in artificial intelligence (AI) applications. These multimodal AI systems have the potential to revolutionize the fields of education, healthcare, and entertainment by enabling individuals to communicate with machines in a natural language that emulates human conversation. The emergence of large language-vision models and the availability of language-vision datasets has greatly improved the performance of many language-vision tasks, such as visual captioning \cite{you2016image,pan2020x,anderson2018bottom}, visual question answering \cite{lei2018tvqa,khan2021towards,lei2020tvqaplus}, and natural language video localization \cite{Hendricks2017LocalizingMI,chen2019localizing}. Recently, video question-answering (VidQA) task has gained attention from natural language processing (NLP) and computer vision (CV) communities as it involves the comprehension of both the video content and language to provide the answer. With the advent of capabilities of large language and vision models, a great level of interaction between machines and humans is possible; however, the majority of the existing works \cite{lei2018tvqa,lei2020tvqaplus,mun2017marioqa} in video question answering focus on identifying the correct natural language answer from the multiple choice answers. Predicting the correct natural language answer may not accurately reflect the real-world scenario, where individuals interact with the machines through natural language queries and seek the appropriate visual segments from the videos, that address their queries. Natural language video localization (NLVL) is one such language-vision understanding task whose goal is to semantically identify a temporal segment within an untrimmed video that is semantically aligned to a language query. Due to its applications in various downstream tasks, such as video retrieval \cite{francis2017natural}, relation detection \cite{rodriguez2021dori}, and visual question answering \cite{lei2018tvqa}, there has been a growing research interest in this direction. However, much of the advancement in VidQA and NLVL is only confined to open-domain, partially due to the availability of large-scale datasets. A closed domain, such as the medical and healthcare domain, where there is a multitude of applications of VidQA and NLVL tasks, remains unexplored. 

\begin{figure}
    \centering
    \includegraphics[width=0.8\linewidth]{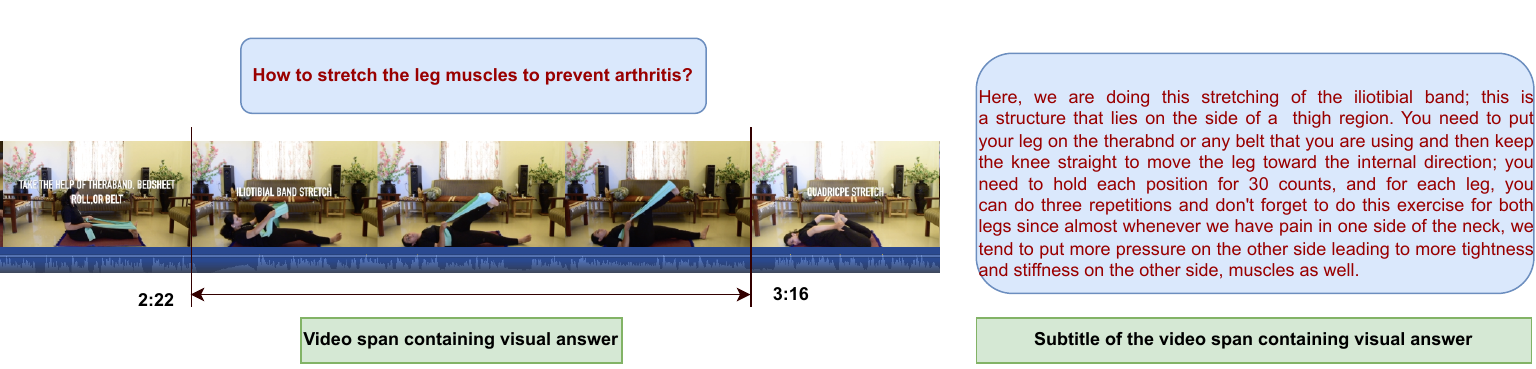}
    \caption{An example of a health-related question and its video answer
 from the video.}
    \label{fig:into}
\end{figure}

Recently, the popularity of online health information searches and the surge in the availability of diverse online educational videos have led healthcare consumers to prefer medical videos over text-based information. This preference arises because medical videos can provide clear visual demonstrations of medical procedures, conditions, and anatomical structures. This visual clarity makes it easier for viewers to grasp complex medical concepts and educate themselves. Moreover, medical videos can depict real-life scenarios and patient interactions, offering viewers a more realistic understanding of medical practices and situations than textual descriptions alone can offer. Consider a health-related question, ``\textit{how to stretch the leg muscles to prevent arthritis?}" (\textit{cf.} Fig. \ref{fig:into}); the textual answer to this question may not be appropriate to act upon for a consumer with limited medical understanding. In this case, a short visual answer will be helpful for the consumer to follow as it offers visual assistance in the form of a step-by-step demonstration. In order to provide visual answers to the consumer's question, a multimodal system should be capable of identifying the relevant videos and locating the appropriate segments from the videos, which can be considered as the audio-visual answer. Providing audio-visual answers from videos can cater to a wider audience, including those with reading difficulties or language barriers. 
Motivated by this, in this work, we focus on the task of medical visual answer localization, with the goal of locating visual answers to medical/healthcare-related questions. In the literature, there is an attempt \cite{gupta2023dataset} to create the dataset; however, the dataset size is small, containing $3,010$ question, answer, and video triplets from $899$ videos. The small size of the dataset hinders the development of sophisticated neural-based approaches, which leads to sub-optimal performance on medical visual answer localization task. To address these issues, we first propose a pipelined approach to automatically create large-scale datasets for the medical visual answer localization task. With this approach, we create two large-scale datasets: HealthVidQA-CRF and HealthVidQA-Prompt, having $23,434$ and $52,711$ question-answer-video triplets, respectively. Second, we propose monomodal and multimodal approaches utilizing the created datasets. The proposed approaches achieve substantial performance improvement on multiple evaluation metrics for the medical visual answer localization task. We summarize the contributions of this work as follows:
\begin{enumerate}
    \item We proposed a three-stage pipelined approach to automatically generate the datasets for the medical visual answer localization task. The performance of each component of this pipeline is validated with human evaluation. The human evaluation reveals that the approaches used in the pipeline are effective and can be used to generate high-quality datasets for the visual answer localization task.
    \item We created two large-scale datasets, HealthVidQA-CRF and HealthVidQA-Prompt, for the task of medical visual answer localization. The former consists of $23,434$ question-answer-video triplets from $11,708$ medical videos, while the latter has $52,711$  triplets generated from $13,990$ medical videos.
    \item We proposed an effective Cycle-Consistent Answer Localization (CCAL) approach, which outperforms the existing approaches on benchmark datasets. Later, we integrated the visual information from multiple visual encoders into the CCAL framework and performed the experiments that showed that the created HealthVidQA-CRF dataset can be used to achieve better performances with multimodal approaches.
    \item We performed a detailed analysis of the results and highlighted the effects of the created datasets in model training, as well as the role of the visual features in improving the performance and benchmarking of the created datasets with monomodal and multimodal approaches. We believe that the created datasets will be used to effectively utilize the pre-trained language vision models to further improve the performance of medical visual answer localization task.
\end{enumerate}

\section{Related Work} \label{sec:related}

\subsection{Video Question Answering}
Video Question Answering is an emerging and challenging task that requires the understanding of video, language, and their interaction to correctly predict the answer to the question. In the literature, the VidQA task involves selecting the correct natural language answer from the multiple-choice natural language answers by performing visual and temporal reasoning on the video.
\subsubsection{Datasets}
\citet{zhu2017uncovering} created the Fill in the blanks dataset that contains questions inferring
the past, describing the present, and predicting the future. The answer has to be filled in from the multiple available choices. This dataset is focused on the Cooking domain. \citet{tapaswi2016movieqa} released the MovieQA dataset containing the multiple-choice answers (MCA) for the human-annotated questions. They utilized the $1075$ videos from the Movies domain and created $14,944$ questions. MarioQA \cite{mun2017marioqa} is a large-scale dataset from the Game domain, where the questions are automatically generated using a set of pre-defined templates. The main goal of this dataset is to answer event-centric questions by temporal reasoning. \citet{jang2017tgif} created the TGIF-QA dataset for open-domain video question answering. The questions are automatically generated, and answers are formulated as multiple-choice answers. \citet{kim2017deepstory} introduced the Pororo-QA dataset for video story question answering. They used the Amazon Mechanical Turk (AMT) service to annotate the MCA-based dataset in the Cartoon domain. YouTube2Text-QA \cite{ye2017video} is another video question-answering dataset constructed from the YouTube2Text video caption dataset. The questions are automatically generated following the approach discussed in \cite{antol2015vqa}, and multiple-choice answers are provided. TVQA dataset was introduced by \citet{lei2018tvqa} by taking the videos ($21,793$) from TV serials and using the AMT service to create the MCA-based questions. Later, they extended the dataset with bounding boxes, linking depicted objects to visual concepts in questions and answers, and released the TVQA+ dataset \cite{lei2019tvqa+}.
\subsubsection{Methods} \citet{tu2014joint} proposed a method based on a joint parse graph of videos and their textual description. A parse graph is constructed to model the various information, such as spatial information, temporal information, and the causalities between events and objects. \citet{zhu2017uncovering} introduced the encoder-decoder approach using Recurrent Neural Networks (RNNs) to learn temporal structures of videos and introduce a dual-channel ranking loss to answer multiple-choice questions. They used GoogLeNet \cite{szegedy2015going} to extract the frame-level features from the video. The textual features are extracted from Skip-thought vectors \cite{kiros2015skip}. \citet{tapaswi2016movieqa} proposed two techniques for VidQA. The first technique called Searching Student with a Convolutional Brain (SSCB), constructs a neural network that assists in answering the question by learning the similarity function. The second method uses a memory network to learn the similarity between question candidate answers. \citet{chu2018forgettable} proposed two mechanisms, the re-watching, and the re-reading mechanisms, and then combined them into an effective forgettable-watcher model. \citet{zhao2018open} introduced adaptive hierarchical reinforced encoder-decoder network learning for open-ended VidQA. They propose the adaptive hierarchical encoder network to learn the joint representation of the long-form video contents according to the question. They also develop the reinforced decoder network to generate the natural language answer. Generative adversarial network (GAN) \cite{goodfellow2014generative} is also utilized in the works \cite{yuan2020adversarial,zhao2020open} for video question answering.

\subsection{Natural Language Video Localization}
Natural language video localization requires modeling the cross-modality interactions between video and natural language (often questions) to retrieve the relevant video segments from the video. \citet{Hendricks2017LocalizingMI} proposed a Moment Context Network (MCN) that learns a shared embedding for video temporal context features and LSTM language features. The proposed video temporal context features integrate local and global video features and temporal endpoint features, which indicate when a moment occurs in a video. They further extended their work and proposed the Moment Localization with Latent Context (MLLC) \cite{hendricks2018localizing}, which models video context as a latent variable. It provides the flexibility to the model to attend to different video contexts conditioned on the specific query/video pair. \citet{Liu2018AMR}  develop a cross-modal retrieval technique to retrieve moments from the video responding to a given query. To align the moment candidates and the input query, they develop a memory attention model to dynamically compute the visual attention over the query and its context information. Other works also exploit the temporal relationship to tackle NLVL problems\cite{Liu2018TemporalMN,zhang2019exploiting,Liu2018CML}. \citet{zhang-etal-2020-span} propose span-based question answering to solve NLVL task. They propose the VSLNet approach based on a query-guided highlighting strategy to search for the target moment within a highlighted region. Later, they extend VSLNet to VSLNet-L, which employs a multi-scale split-and-concatenation approach. The VSLNet-L \cite{zhang2021natural} model first untrimmed video into shorter clip segments. Following that, it identified the clip segment containing the target moment, downplaying the significance of the remaining segments \cite{mavroudi2023learning}.
\subsection{Medical Visual Answer Localization}
\citet{gupta2023dataset}, first introduced the task of medical visual answer localization (MVAL) and created the MedVidQA dataset having $3,010$ question-answer-video triplets. They also benchmark the dataset with baseline systems following the work of \citet{zhang-etal-2020-span}. Later, they utilized the dataset and organized the shared task \cite{gupta-demner-fushman-2022-overview} of retrieving the answer segments from the video against the health-related question. The majority of the participants \cite{gupta-demner-fushman-2022-overview} utilized the pre-trained language models \cite{beltagy2020longformer,zaheer2020big,choromanski2020rethinking} to solve the MVAL task as a reading comprehension problem \cite{rajpurkar2016squad}. \citet{li2022towards} attempt to solve the MVAL problem by introducing the visual highlight prompts into the pre-trained language model (PLM) for enhancing the joint semantic representations of subtitles and video frames. \cite{li2023learning} introduce the Cross-Modal Contrastive Global-Span (CCGS) method for the video corpus visual answer localization task. This method leverages the global-span matrix to jointly train the video corpus retrieval and visual answer localization subtasks. As an alternative, this study focuses on introducing new large-scale medical visual answer localization datasets and proposing approaches to effectively solve the MVAL problem.

\section{Generating Large-scale Medical Instructional VideoQA Dataset} \label{sec:data_creation}
This section describes our methodology for automatically creating a comprehensive medical instructional visual answer localization dataset.  We discuss the strategy to generate the video, question, and answer triplets. Next, we outline the analysis and human evaluation of the generated dataset. 
\subsection{Generating Video-question-answer Triplets} This subsection deals with choosing the medical instructional videos and determining the visual segments in videos that could serve as the visual answer to the medical or health-related questions.  Furthermore, we describe our methodology for generating medical instructional questions from the subtitles of the videos.  
\subsubsection{Selecting Medical Instructional Videos} In the first step of generating video-question-answer triplets, we aim to select the medical videos that can be used in subsequent steps of the dataset creation. We leverage the videos from `\textit{Personal Care and Style},' `\textit{Health},' and `\textit{Sports and Fitness}' categories within the HowTo100M \cite{miech2019howto100m} dataset. The HowTo100M dataset has $1.22$M videos of different categories, out of which `\textit{Personal Care and Style},' `\textit{Health},' and `\textit{Sports and Fitness}' categories contain $16K$, $15K$ and $15K$ videos, respectively. To construct, a medical instructional visual answer localization dataset, \textbf{(1)} a video should describe a health-related topic, such as diseases, medical conditions, symptoms, drugs, treatments, medical exams, and procedures, etc., \textbf{(2)} video should clearly demonstrate a step-by-step medical procedure providing enough details to reproduce the procedure and achieve the desirable results. We observed that the automatic video category labeled in the HowTo100M dataset was not always accurate. Moreover, some of the videos were also not instructional in nature, which hindered the development of the medical instructional visual answer localization dataset. To address this issue, we utilized the video classification MedVidCL dataset from \citet{gupta2023dataset}, which has a total of $4,217$ training videos annotated for `\textit{Medical Instructional},' `\textit{Medical Non-instructional}' and `\textit{Non-Medical}.' Following, \cite{gupta2023dataset}, we fine-tune the BigBird$_{\text{Base}}$ \cite{zaheer2020big} model on the training set of MedVidCL dataset by extracting the video subtitles using the \texttt{pytube}\footnote{\url{https://pytube.io/en/latest/}} library. We evaluated the performance of the fine-tuned BigBird$_{\text{Base}}$ model on the test set of MedVidCL and recorded the F1-score of $94.28\%$ on detecting the medical instructional videos. The fine-tuned video classification model was used to label the subset (`\textit{Personal Care and Style},' `\textit{Health},' and `\textit{Sports and Fitness}' categories videos) of the HowTo100M videos into medical instructional videos. This process yielded $15,664$ medical instructional videos that we used in the subsequent steps of the dataset creation.
\subsubsection{Detecting Visual Answer Segments} \label{sec:answer-segments}
Given a medical instructional video $V$ having raw subtitle/caption list $C_V=\{c_1, c_2, \ldots, c_m\}$ and corresponding time-stamps $T_V=\{t_1, t_2, \ldots, t_m\}$  of length $m$, where $c_i$ is the $i^{th}$ span of subtitle having the time stamps $t_i$. The issue with the raw subtitles is that they are not segmented and often overlap with the previous subtitles. To alleviate this issue, we concatenated the subtitle list $C_V$ and formed a sequence of words $W_V=\{w_1, w_2, \ldots, w_{|W_V|}\}$. In the next step, we hypothesize that the subtitle describing a visual answer in the video corresponds to a particular topic. Towards this, we aim to obtain the topic-aware segment from the $W_V$, and utilize the DeepSegment~\footnote{\url{https://github.com/notAI-tech/deepsegment/tree/master}} model to segment the $W_V$ into $k$ topic-aware segments $S_V=\{s_1, s_2, \ldots, s_k\}$, next, we align the time-stamps $T_V$ to the topic-aware segments and obtain the aligned time-stamps $\hat{T}_V=\{t_1, t_2, \ldots, t_k\}$. With the topic-aware segments $S_V$ and corresponding aligned time-stamps $\hat{T}_V$ of the video $V$, Our goal is to identify topics that describe visual answers in the videos and subsequently divide the input sequence into contiguous segments representing distinct topics. We propose two approaches for detecting visual segments: \textbf{(1)} XLNet-CRF Model,  and \textbf{(2)} XLNet-Prompt Model. In the XLNet-CRF model, we consider the problem of visual answer segment detection as the sequence labeling problem, where the beginning of the segment is marked as `B-Seg,' intermediate as `I-Seg,' and other segments as `O'. 
The XLNet-Prompt model is based on predicting the masked word using a language model. We designed prompts and mapped the masked word with the appropriate label, which signifies the beginning, intermediate, and other segments.
Below, we outline both approaches to detect the visual answer segments from the subtitles of the video:
\paragraph{XLNet-CRF Model:}
This approach utilizes the pre-trained \xlnet{} \cite{yang2019xlnet} model to encode the segments and make the decision by using the conditional random field \cite{lafferty2001conditional} based tagger to tag the boundaries of the visual segments. Considering the segments $S_V=\{s_1, s_2, \ldots, s_k\}$ from video $V$. We performed the following operations to tag the boundaries of the visual segments that can be considered visual answers to instructional questions. 
\begin{enumerate}
    \item \textbf{Segment Encoding}: This module takes the segments as input and processes them, and returns the encoded representation of the segment. Particularly, we obtain the hidden state representation $H_i$ of each segment $s_i$ of token length $|s_i|$ using the last layer of \xlnet{} model. Thereafter, we choose the last token, hidden state representation, as the segment representation.
   Formally:
    \begin{equation} \label{eq:xlnet-hidden}
\begin{split}
 H_1,  H_2, \ldots, H_k &= \texttt{XLNet}(s_1, s_2, \ldots, s_k), ~ \text{where}~ H_i\in \mathcal{R}^{|s_i| \times d}\\
   h_1,  h_2, \ldots, h_k &= \texttt{Select}([H_1,  H_2, \ldots, H_k], \text{last}), ~ \text{where}~ h_i\in \mathcal{R}^{d}
\end{split}
\end{equation}
Where \texttt{Select} (; last) is an operation that chooses the last token hidden state representation, and $d$ is the dimension of the \xlnet{} hidden state representation.
    \item \textbf{Segment Sequence Processing: } This step aims to process the encoded segments in the form of a sequence. However, the hidden state representations $\{h\}_{i=1}^{i=k}$ of the segments obtained in the previous step do not hold the inherent notion of the segment order. To tackle this, we first introduce the positional information of the segments in the form of positional embedding. Specifically, we augmented the positional embedding $p_i$ into the segment representation $h_i$ to obtain the position-aware segment representation $h_i^*$.    Formally: 
     \begin{equation} \label{eq:positional-emb}
\begin{split}
 h_1^*,  h_2^*, \ldots, h_k^{*} &= 
(h_1+p_1), (h_2+p_2),\ldots, (h_k+p_k),  ~ \text{where}~ p_i\in \mathcal{R}^{d}
\end{split}
\end{equation}
To process the segment representation $\{h^*\}_{i=1}^{i=k}$, we employed a \transformerenc{} layer \cite{devlin-etal-2019-bert}, which utilizes the attention mechanism \cite{vaswani2017attention} to transform segment hidden states into rich and context-aware segment representations. 
 \begin{equation} \label{eq:transformer}
\begin{split}
 u_1,  u_2, \ldots, u_k &= \transformerenc{}(h_1^*,  h_2^*, \ldots, h_k^{*}),  ~ \text{where}~ u_i\in \mathcal{R}^{d}
\end{split}
\end{equation}
Given the context-aware segment representations $U=\{u\}_{i=1}^{i=k} \in \mathcal{R}^{k \times d}$, we use a feed-forward network to project each segment representation $u_i$ into $c$-dimensional (exhibits the B-Seg, I-Seg and
O tags) score $l$ as follows:
 \begin{equation} \label{eq:classifier}
\begin{split}
l &=  \mathbf{W}U +\mathbf{b},  ~ \text{where}~ \mathbf{W}\in \mathcal{R}^{d \times c}, \mathbf{b} \in \mathcal{R}^{c}, p\in \mathcal{R}^{k \times c} 
\end{split}
\end{equation}
      \item \textbf{CRF-based Segment Tagging: } The output score $l$ obtained in the sequence processing step does not account for the dependencies across output labels. Segment labeling is one such task in which the label assigned to the preceding segment plays a crucial role in guiding the current segment to make accurate predictions. To achieve this, we utilized the CRF, which models the tagging decisions jointly. More formally, given the segments $S_V$ and prediction $y=\{y_1, y_2, \ldots, y_k\}$, the score $\mathcal{S}$ is computed as follows:
       \begin{equation} \label{eq:crf}
\begin{split}
\mathcal{S}(S_V, y) &=  \sum_{i=2}^{k}M[y_{i-1}][y_i] + \sum_{i=1}^{k}l_i[y_{i}]
\end{split}
\end{equation}
where $M$ is the matrix that contains the transition score between two subsequent labels. To train the network, the model maximizes the log probability of the correct segment sequence. In the testing phase, a sequence of predicted labels $y^*$ that maximize the score $\mathcal{S}$ is chosen as the final segment label sequence.
\end{enumerate}
\paragraph{XLNet-Prompt Model:} Inspired by the success of prompting \cite{liu2023pre} that aims to bridge the gap between pre-training and fine-tuning of the language model, we also explore prompt-based fine-tuning to tag the boundaries of the visual segments.
\begin{enumerate}
    \item \textbf{Prompt Tuning: } For the task of detecting visual segments, we develop prompts, which are a set of template $T(;)$ and label words $\mathcal{V}$. For each segment $s_i \in S_V$, we apply a prompt template and convert $s_i$ into prompt input $s_i^p$ for \xlnet{} model. The prompt template usually has a \texttt{[MASK]} token, which needs to be filled by a label word $v \in \mathcal{V}$. We fed the prompt input $s_i^p$ into the \xlnet{} model and computed the hidden state representation of $h_i^{ \texttt{[MASK]} }$. Thereafter, we compute the probability that label $v$ can fill the \texttt{[MASK]} token. Formally,
      \begin{equation} \label{eq:xlnet-prompt}
\begin{split}
 s_1^p,  s_2^p, \ldots, s_k^p &= T(s_1, s_2, \ldots, s_k), \\
 H_1,  H_2, \ldots, H_k &= \texttt{XLNet}(s_1^p,  s_2^p, \ldots, s_k^p), ~ \text{where}~ H_i\in \mathcal{R}^{|s_i| \times d}\\
   h_1^{ \texttt{[MASK]} },  h_2^{ \texttt{[MASK]} }, \ldots, h_k^{ \texttt{[MASK]} } &= \texttt{Select}([H_1,  H_2, \ldots, H_k], \texttt{[MASK]} ), ~ \text{where}~ h_i^{ \texttt{[MASK]} }\in \mathcal{R}^{d} \\
    p(\texttt{[MASK]}=v| s_i^p) &= \frac{exp(h_i^v.h_i^{ \texttt{[MASK]} })}{\sum_{\hat{v} \in \mathcal{V}}exp(h_i^{\hat{v}}.h_i^{ \texttt{[MASK]} })}, ~ \forall v \in \mathcal{V}, ~ \text{where}~ h_i^v\in \mathcal{R}^{d}\\
\end{split}
\end{equation}
Finally, we map the segment labels `B-Seg,' `I-Seg,' and `O' to the label words $\mathcal{V}$ to obtain the segment label. The mapping function to map the class into the label words is called a `verbalizer' in the literature \cite{liu2023pre}.  
    \item \textbf{Template and Label Words:} We performed a series of experiments with multiple templates for the visual segment detection task.  With the supervised data, the pre-trained language model can be fine-tuned to maximize the log-likelihood of the correct segment labels. We have outlined all the templates used in this work in Table \ref{tab:prompt-perfomance}. The label words used for the label to token mapping are as follows: `B-Seg': `\textit{first},' `I-Seg': `\textit{next},' \\ `O': `\textit{other}.'

\begin{table}[]
    \centering
 \resizebox{\textwidth}{!}{%
\begin{tabular}{l|l|ccc|ccc}
\hline
\multirow{2}{*}{\textbf{Id}} & \multirow{2}{*}{\textbf{Template}}                                               & \multicolumn{3}{c|}{\textbf{MedVidQA}} & \multicolumn{3}{c}{\textbf{MVAL}} \\ \cline{3-8} 
& & \multicolumn{1}{c}{\textbf{\begin{tabular}[c]{@{}c@{}}F1-Score\\ (w=1)\end{tabular}}} & \multicolumn{1}{c}{\textbf{\begin{tabular}[c]{@{}c@{}}F1-Score\\ (w=2)\end{tabular}}} & \textbf{\begin{tabular}[c]{@{}c@{}}F1-Score\\ (w=3)\end{tabular}} & \multicolumn{1}{c}{\textbf{\begin{tabular}[c]{@{}c@{}}F1-Score\\ (w=1)\end{tabular}}} & \multicolumn{1}{c}{\textbf{\begin{tabular}[c]{@{}c@{}}F1-Score\\ (w=2)\end{tabular}}} & \textbf{\begin{tabular}[c]{@{}c@{}}F1-Score\\ (w=3)\end{tabular}} \\ \hline
\hline

1           & \mask{} \textless{}SEG\textgreater{}   & 0.4014 & 0.5016 &  0.5701
&  0.3747 & 0.4769 & 0.5380  
\\ 
2           & \mask{} \sep{} \textless{}SEG\textgreater{}                               &    0.3152 &  0.4535 &  0.5130 &           0.3463 & 0.4361 &  0.4918                                                                                \\ 
3           & \textless{}SEG\textgreater \sep{} \mask{}                                  &                 0.3142 &    0.4226 &    0.5098 &      0.2985 &  0.4152 &      0.5401                                                                 \\ 
4           & This is the \mask{} step where \textless{}SEG\textgreater{}      &     0.2983 & 0.4275 &   0.4885 & 0.4225  & 0.4945 & 0.5583                                                                                           \\
5           & This is the \mask{} step where\sep{} \textless{}SEG\textgreater                      &                     0.3283 &     0.4740 & 0.5279     & 0.3821 &  0.4864 &                  0.5596                                                    \\ 
6           & This is the \mask{} step  \textless{}SEG\textgreater{}           &                           0.3381 &  0.4642 &  0.5212 &                                        0.4064 &   0.5418 &  0.5905                          \\ 
7           & This is the \mask{} step \sep{}\textless{}SEG\textgreater{} &     0.3217 & 0.4896 &  0.5710   & 0.3595 & 0.5064 &0.5807                                                                                          \\ 
8           & \mask{} I am going to \textless{}SEG\textgreater{}               &   0.3042 &   0.4474 &  0.5124   & 0.3825 &                         0.4854&                              0.5602                                      \\ 
9           & \mask{} I am going to \sep{} \textless{}SEG\textgreater{}     &                              0.2861 &     0.4536 &  0.5343 & 0.2766 &    0.4188 &   0.5051                                                          \\ \hline \hline
\end{tabular}%
}
\caption{Performance comparison of the prompt-based segment detection approach on the test set of the MedVidQA and MVAL datasets.}
  \label{tab:prompt-perfomance}
    
\end{table}
\begin{table}[]
    \centering

\resizebox{0.85\textwidth}{!}{%
\begin{tabular}{l|ccc|ccc}
\hline
\multirow{2}{*}{\textbf{Models}} & \multicolumn{3}{c|}{\textbf{MedVidQA}} & \multicolumn{3}{c}{\textbf{MVAL}} \\ \cline{2-7} 
 & \multicolumn{1}{c}{\textbf{\begin{tabular}[c]{@{}c@{}}F1-Score\\ (w=1)\end{tabular}}} & \multicolumn{1}{c}{\textbf{\begin{tabular}[c]{@{}c@{}}F1-Score\\ (w=2)\end{tabular}}} & \textbf{\begin{tabular}[c]{@{}c@{}}F1-Score\\ (w=3)\end{tabular}} & \multicolumn{1}{c}{\textbf{\begin{tabular}[c]{@{}c@{}}F1-Score\\ (w=1)\end{tabular}}} & \multicolumn{1}{c}{\textbf{\begin{tabular}[c]{@{}c@{}}F1-Score\\ (w=2)\end{tabular}}} & \textbf{\begin{tabular}[c]{@{}c@{}}F1-Score\\ (w=3)\end{tabular}} \\ \hline
\hline
\begin{tabular}[c]{@{}l@{}}BERT-CRF \cite{devlin-etal-2019-bert} \end{tabular} & 0.3671 & 0.5191 & 0.6184 & 0.2923& 0.4725 & 0.5712 \\ 
\begin{tabular}[c]{@{}l@{}}ALBERT-CRF \cite{Lan2020ALBERT} \end{tabular} & 0.2981 & 0.4417 & 0.5854 &0.2475 &0.3997 &0.4951 \\ 
\begin{tabular}[c]{@{}l@{}}ELECTRA-CRF \cite{clark2020electra}\end{tabular} & 0.4110 & \textbf{0.5671} & 0.6225&0.3028 &0.4167 &0.5667 \\ 
\begin{tabular}[c]{@{}l@{}}RoBERTa-CRF \cite{liu2019roberta}\end{tabular} & 0.3860 & 0.5643 & \textbf{0.6462} & 0.3192& 0.3974& 0.5097\\ 
\begin{tabular}[c]{@{}l@{}}XLNet-CRF \cite{yang2019xlnet}\end{tabular} & \textbf{0.4183} & 0.5256 & 0.6112 & \textbf{0.3216} &\textbf{0.4904} & \textbf{0.5901} \\ \hline \hline 
\end{tabular}%
}
  \caption{Performance comparison of the CRF-based segment sequence labeling on the test set of MedVidQA and MVAL datasets.}
  \label{tab:crf-perfomance}
    
\end{table}

\end{enumerate}
\paragraph{Training and Evaluation of Models:} To train the models, we utilize the MedVidQA dataset \cite{gupta2023dataset}, where video, question, and visual answer segment's beginning and end time-stamps are provided. We followed the strategy discussed above to transform the video subtitle into segments and marked each segment as start (`B-Seg'), intermediate (`I-Seg'), or other (`O') segments. The MedVidQA dataset has a total of $2,710$, $1,45$, and $1,55$ visual segments in training, validation, and test sets, respectively. We train XLNet-CRF and XLNet-Prompt models on the training set of the MedVidQA dataset, tune the hyper-parameters on the validation set, and evaluate the performance on the test set of the MedVidQA and MVAL task ($1,53$ visual segments) of MedVidQA shared task \cite{gupta-demner-fushman-2022-overview} datasets. We evaluated the performance of the approaches in terms of F1-score. While collating the subtitles and segmenting them based on DeepSegment, we observed that some text from the segment may fall in the previous and next segments. Since we are labeling the segments based on their time stamps that is mapped from the raw subtitles, it may be unfair to evaluate the performance with the strict boundaries of the segments. Therefore, while computing the true positive for the segment label, we relax it via a window of $w \in \{1,2,3\}$. Given a window of size $w$, we consider the predicted segment a correct segment if It is off by $w$ segments to either left or right in the segment sequence. Following this, we evaluated the XLNet-CRF models with multiple competitive approaches and reported the performance in Table \ref{tab:crf-perfomance}. XLNet-CRF model achieved the F1-score of $0.418$ in the most restricted evaluation setting with $w=1$ for the MedVidQA dataset and outperformed the competitive models on all the window settings ($w\in \{1,2,3\}$) for MVAL test dataset. The performance of the \xlnet{} model inspires us to assess its capability in prompt tuning setup too. The detailed performance comparison of the XLNet-Prompt model on multiple templates is reported in Table \ref{tab:crf-perfomance}. 
\begin{table}[]
\centering
\resizebox{0.95\textwidth}{!}{%
\begin{tabular}{ll|cccccccc}
\hline
\textbf{Modality}&\textbf{Models} & \textbf{BLEU-1} & \textbf{BLEU-2} & \textbf{BLEU-3} & \textbf{BLEU-4} & \textbf{ROUGE-1} & \textbf{ROUGE-2} & \textbf{ROUGE-L} & \textbf{BERTScore} \\ \hline \hline
Vision & Enc-Dec \cite{vaswani2017attention} & 32.48 & 17.81 & 9.980 & 6.670 & 30.2894 & 12.3377 & 28.8729 & 64.12 \\ 
Language +Vision & UniVL \cite{luo2020univl} & 40.39 & 24.98 & 16.18 & 11.40 & 41.7152& 19.2782 & 39.2738 & 68.90 \\ 
Language &PEGASUS \cite{zhang2020pegasus}& 42.46 & 28.87 & 20.06 & 14.70 & 47.6677 & 25.4939 & 45.1603 & 71.49 \\ 
Language& BART \cite{lewis2019bart} & 43.06 & 28.53 & 19.22 & 13.71 & 44.9675 & 23.5237 & 42.5070 & 69.59 \\ 
Language & T5 \cite{raffel2020exploring} & 43.97 & 30.42 & 20.99 & 15.43 & 47.6023 & 25.0923 & 44.6815 & 70.23 \\ \hline \hline
\end{tabular}%
}
\caption{Performance comparison of the multiple language and vision models on medical instructional question generation on the test set of MedVidQA.}
\label{tab:qg-results}
\end{table}
\subsubsection{Generating Medical Instructional Questions} Given the visual answer segments $S_V^a=\{s_1, s_2, \ldots, s_r\}$ and aligned timestamps $\hat{T}_V^a=\{t_1, t_2, \ldots, t_r\}$  of length $r$ of video $V$ detected using the approaches discussed in Section \ref{sec:answer-segments}, the goal of this component is to generate instructional questions focusing on medical or health-related topics. Toward this, we built parameterized question generation models and optimized the parameters using the segment-question pairs available in the MedVidQA dataset. We explored monomodal and multimodal approaches to generate instructional questions by utilizing the respective modality from the video. Considering an answer segment $s_i \in S_V^a$ and respective timestamp $t_i = (t_i^{s}, t_i^{e})$, where $s$ and $e$ denote the start and end timestamp, in the vision-based monomodal approach, we consider the frames $f_i=\{f_i^s, f_i^{s+1}, \ldots, f_i^{e}\}$ spanning between $t_i^{s}$ and $t_i^{e}$ in the video $V$ and train a Transformer-based encoder-decoder (Enc-Dec) model to generate the question. For the language-based monomodal approach, we collate all the sub-segments from $s_i$ and form the sequence $s_i=\{w_i^1, w_i^2, \ldots, w_i^{|s_i|}\}$ and fine-tune the pre-trained language models (PEGASUS, BART, and T5) on the MedVidQA dataset to generate the question. For the multimodal experiment, we fine-tune the UniVL \cite{luo2020univl} pre-trained language-vision model by using the frames $f_i$ and $s_i$ from the MedVidQA dataset. We have reported the performance from monomodal and multimodal approaches in terms of BLEU \cite{papineni2002bleu}, ROUGE \cite{lin2004rouge}, and BERTScore \cite{zhang2019bertscore} in Table \ref{tab:qg-results}. It can be observed from Table \ref{tab:qg-results} that the T5 model outperformed other approaches on the BLEU-4 metric, which has been a standard evaluation metric for question generation \cite{du2017learning,gupta2020reinforced,dong2019unified} task. Therefore, we consider the T5-based question generator to generate the instructional question.
\subsection{HealthVidQA: large-scale  Medical Instructional VideoQA dataset}
We have utilized the procedure outlined earlier and created two large-scale \textbf{Health} \textbf{Vid}eo \textbf{Q}uestion \textbf{A}nswering (HealthVidQA) datasets: HealthVidQA-CRF and HealthVidQA-Prompt. In the HealthVidQA-CRF, we used the XLNet-CRF approach to identify the visual segments and used the T5-based question generation approach to generate instructional questions.  The second dataset (HealthVidQA-Prompt) used the prompt-tuning approach to identify the visual segment. Similar to the former dataset, T5-based question generation was used to generate the instructional questions. We present the dataset analysis and human evaluation in the following subsections:
\begin{figure}[]
\begin{minipage}{.55\textwidth}
  \centering
  \resizebox{\textwidth}{!}{%
\begin{tabular}{lcc}
\hline
\textbf{Models} & \textbf{\begin{tabular}[c]{@{}c@{}}HealthVidQA-CRF\end{tabular}} & \textbf{\begin{tabular}[c]{@{}c@{}}HealthVidQA-Prompt\end{tabular}} \\ \hline \hline
\# Question-Answer & 23,434 & 52,771 \\
\# Videos & 11708 & 13,990 \\ 
Mean Question Length & 9.58 & 9.72 \\ 
Max Question Length & 19 & 19 \\ 
Min Question Length & 5 & 5 \\ 
Mean Subtitle Length & 249 & 114.14 \\ 
Max Subtitle Length & 5926 & 2017 \\ 
Min Subtitle Length & 20 & 20 \\ 
Mean Visual Answer Length& 73.33 & 33.88 \\ 
Max Visual Answer Length & 1,110 & 1,763 \\
Min Visual Answer Length & 5 & 5 \\ \hline \hline
\end{tabular}%
}
\captionof{table}{Detailed questions, videos, and visual answers statistics of the datasets created using XLNet-CRF and XLNet-Prompt approaches. Question and subtitle length are measured in terms of words, whereas visual answers are measured in seconds.}
\label{tab:dataset-stats}

\end{minipage}
\centering
\quad
\begin{minipage}{.4\textwidth}
  \centering
\resizebox{\textwidth}{!}{%
\begin{tabular}{ccclccc}
\cline{1-7}
\multicolumn{3}{c}{\textbf{\begin{tabular}[c]{@{}c@{}}HealthVidQA-CRF\end{tabular}}} & \multicolumn{1}{l|}{} & \multicolumn{3}{c}{\textbf{\begin{tabular}[c]{@{}c@{}}HealthVidQA-Prompt\end{tabular}}} \\ \hline \hline\\
\multicolumn{7}{c}{\textbf{Medical-Instructional Videos}} \\ \hline
\multicolumn{1}{c}{\textbf{Yes}} & \multicolumn{2}{c}{\textbf{No}} & \multicolumn{1}{l|}{} & \multicolumn{1}{c}{\textbf{Yes}} & \multicolumn{2}{c}{\textbf{No}} \\ 
\multicolumn{1}{c}{81.61} & \multicolumn{2}{c}{18.39} & \multicolumn{1}{l|}{} & \multicolumn{1}{c}{81.61} & \multicolumn{2}{c}{18.39} \\ \hline  \\
\multicolumn{7}{c}{\textbf{Segment Containing Visual Answer}} \\ \hline
\multicolumn{1}{c}{\textbf{Yes}} & \multicolumn{1}{c}{\textbf{No}} & \multicolumn{1}{c}{\textbf{Partial}} & \multicolumn{1}{l|}{} & \multicolumn{1}{c}{\textbf{Yes}} & \multicolumn{1}{c}{\textbf{No}} & \textbf{Partial} \\ 
\multicolumn{1}{c}{82.04} & \multicolumn{1}{c}{6.59} & \multicolumn{1}{c}{11.38} & \multicolumn{1}{l|}{} & \multicolumn{1}{c}{75.45} & \multicolumn{1}{c}{5.99} & 18.56 \\ \hline \\
\multicolumn{7}{c}{\textbf{Question Generation Assessment}} \\ \hline
\multicolumn{1}{c}{\textbf{Correct}} & \multicolumn{1}{c}{\textbf{Incorrect}} & \multicolumn{1}{c}{\textbf{Partial Correct}} & \multicolumn{1}{l|}{} & \multicolumn{1}{c}{\textbf{Correct}} & \multicolumn{1}{c}{\textbf{Incorrect}} & \textbf{Partial Correct} \\ 
\multicolumn{1}{c}{77.38} & \multicolumn{1}{c}{13.10} & \multicolumn{1}{c}{9.52} & \multicolumn{1}{l|}{} & \multicolumn{1}{c}{62.59} & \multicolumn{1}{c}{35.97} & 1.44 \\ \hline \\
\multicolumn{7}{c}{\textbf{Segment Question Alignment}} \\ \hline
\multicolumn{1}{c}{\textbf{Yes}} & \multicolumn{1}{c}{\textbf{No}} & \multicolumn{1}{c}{\textbf{Partial}} & \multicolumn{1}{l|}{} & \multicolumn{1}{c}{\textbf{Yes}} & \multicolumn{1}{c}{\textbf{No}} & \textbf{Partial} \\
\multicolumn{1}{c}{75.45} & \multicolumn{1}{c}{5.99} & \multicolumn{1}{c}{18.56} & \multicolumn{1}{l|}{} & \multicolumn{1}{c}{46.67} & \multicolumn{1}{c}{27.41} & 25.93 \\ \hline \hline
\end{tabular}%
}
  \captionof{table}{Comparison of the human evaluation scores (on multiple criteria) for the datasets created using XLNet-CRF and XLNet-Prompt approaches.}
  \label{tab:human-eval}
\end{minipage}
\end{figure}
\subsubsection{Dataset Analysis} The HealthVidQA-CRF has $23,434$ video-question-answer triplets from $11,708$ medical videos. Each video has an average of $2$ visual answer segments of a duration of $73.33$ seconds. The minimum and maximum duration of the visual answer segment are $5$ and $1,110$ seconds, respectively. The average generated question length is $9.58$ words with a minimum and maximum length of $5$ and $19$ words, respectively. We have also performed the analysis of the HealthVidQA-Prompt dataset. The HealthVidQA-Prompt dataset has $52,771$ video-question-answer triplets from $13,990$ medical videos. We observed that each video has an average of $3.77$ visual answer segments of a duration of $33.88$ seconds. For the generated questions, the statistics match the HealthVidQA-CRF question. On average generated question length is $9.72$, with a minimum and maximum length of $5$ and $19$, respectively. We have provided detailed statistics of both the created datasets in Table \ref{tab:dataset-stats}. We observe that the HealthVidQA-Prompt dataset has shorter segments, thus a shorter subtitle length, which leads to more visual segments in this dataset. We will evaluate the quality of both the datasets in the next subsection.
\subsubsection{Human Evaluation} We have followed an automated way to create the dataset, which could lead to noisy samples. In the dataset creation process, the possible causes of the errors could be in \textbf{(1)} selecting medical instructional videos, \textbf{(2)} detecting segment containing a visual answer, \textbf{(3)} generating valid instructional questions and \textbf{(4)} alignment of generated question and predicted segment. In order to assess the quality of the dataset, we devise a series of human evaluations to assess the aforementioned causes of the errors. In our human evaluation setups, we randomly choose $308$ samples from the generated datasets, and a total of three annotators judge the questions, segments, and videos to provide the assessment. Our human scores are the agreement scores amongst the annotators.
\begin{itemize}
    \item \textbf{Medical Instructional videos}: Getting the medical instructional questions is the first step of the dataset creation pipeline; therefore, we performed the human evaluation and asked the annotators to mark the videos whether they are medical instructional or not.
    \item \textbf{Segment Containing Visual Answer}: We proposed two approaches, XLNet-CRF and XLNet-Prompt, for detecting the visual segments from the videos. This evaluation assesses whether the predicted segment contains a visual answer to any medical instructional questions. We asked the annotators to mark `\textbf{Yes}' if the segment contains a complete illustration of a particular procedure, `\textbf{No}' if the segment does not contain any illustration, and `\textbf{Partial}' if the segment contains a partial illustration of a particular procedure.
    \item \textbf{Question Generation Assessment}: With this evaluation, we evaluated the quality of the generated question. Towards this, we asked annotators to mark the question as `\textbf{Correct}' if the question is a well-formed and valid instructional question,`\textbf{Partial Correct}' if the question belongs to either well-formed, has minor errors or valid instructional question, and `\textbf{Incorrect}' if the question is either grammatically, semantically or pragmatically incorrect.
    \item \textbf{Segment Question Alignment}: This evaluation assesses whether the visual segment and corresponding generated question are aligned with each other. We asked the annotators to mark the segment-question pair as `\textbf{Yes},' or `\textbf{Partial}' if the segment contains a complete or partial illustration as a visual answer to the generated instructional question, `\textbf{No}' otherwise.

\end{itemize}
We performed the human evaluation on both the created datasets HealthVidQA-CRF and HealthVidQA-Prompt. The detailed human evaluation is depicted in Table \ref{tab:human-eval}. From the human evaluation, we observe that the HealthVidQA-CRF dataset is more accurate compared to the HealthVidQA-Prompt dataset; therefore, we chose HeathVidQA-CRF for benchmarking the medical visual answer localization task. We believe the HealthVidQA-Prompt dataset has some noise, but it can be helpful for training in low-resource settings \cite{fang2016learning}, bootstrapping \cite{papadopoulou2022bootstrapping}, and building a scalable model \cite{cavinato2023explainable}.  
\section{Approaches} \label{sec:approaches}
\subsection{Cycle-Consistent Answer Localization (CCAL)} \label{sec:ccal}
We proposed an approach to effectively locate the visual segments that contain the answer to the given medical instructional question. With the success of the reading comprehension-based approaches \cite{gupta-demner-fushman-2022-overview} for the medical visual answer localization task, we followed the reading comprehension-based approach where the task aims to locate the start and end timestamps in the video where the answer to the medical question is being shown or the explanation is illustrated in the video. 
Inspired by cycle-consistent training\cite{shah2019cycle}, given a medical instructional question $Q$, video $V$ and corresponding subtitle $S$, and visual answer timestamp $T=(T^s, T^e)$, we first employed a text-based reading comprehension model $f$ with parameters $\theta$, which takes question $Q$ and subtitles $S$ and predict the start and end time stamps of the answer $\hat{T}  \leftarrow f(Q, S; \theta)$, where $\hat{T}=(\hat{T}^s, \hat{T}^e$). Using the predicted answer $\hat{T}$ and subtitle $S$ of video $V$, we utilize a question generation model $g$  with parameters $\phi$, and generate question $\hat{Q} \leftarrow g(S, \hat{T}; \phi)$. Our hypothesis is that if the reading comprehension model $f$ predicts the answer $\hat{T}$ correctly for the question $Q$, then the generated question using $\hat{T}$ and subtitle $S$ will be semantically and syntactically similar to $Q$. 
\subsubsection{Reading Comprehension Model} \label{sec:reading-comp-model} Our reading comprehension model deals with the subtitle $S$ of the video $V$, question $Q=\{q_q, q_2, \ldots, q_{|Q|}\}$ to predict the answer span in the subtitle. To effectively encode the longer subtitle $S$, we use the pre-trained Longformer model \cite{beltagy2020longformer} to encode the subtitle $S=\{s_1, s_2, \ldots, s_n\}$ having $n$ subtitles. Specifically, we concatenate all the words from the subtitles and formulate the word sequence $W=\{w_1^1, \ldots w_1^{|s_1|}, \ldots, w_{n}^{1} \ldots w_{n}^{|s_n|}\}$. Following the fine-tuning of the question-answering model, we packed the question $Q$ and subtitle word sequence $W$ to form a single-word sequence $C$.  We fed the word sequence $C$ to the Longformer model and obtained the hidden state $h_i$ for each token $t_i \in C$. Thereafter, a linear layer is employed on the top of the hidden state to compute the span start logit and span end logit. We predict the start and end positions of the tokens; thereafter, we map these tokens to their corresponding subtitle and extract their time stamp. We call the predicted start and end positions of the time stamp of the answer as $\hat{T}=(\hat{T}^s, \hat{T}^e)$. We train the network by maximizing the log-likelihoods of the correct start and end positions of the answer. We denote the loss function of the network as $\mathcal{L}_f(T, \hat{T})$\footnote{While computing the loss for the reading comprehension model, we calculate the cross-entropy loss between expected and predicted positions of the answer's starting and ending words within the word sequence denoted as $W$.}.
\subsubsection{Question Generation Model} \label{sec:qg-model}With the predicted start and end positions of the time stamp of the answer as $\hat{T}=(\hat{T}^s, \hat{T}^e)$, we map them to their corresponding word sequence $\hat{W}$ in $W$ and pass the word sequence $\hat{W}$ to  BART \cite{lewis2019bart} to generate the instructional question. We call the generated question $\hat{Q}$. The question model is trained by maximizing the log-likelihood of the correct question $Q$. We denote the loss function of the network as $\mathcal{L}_g(Q, \hat{Q})$. 
We train our proposed CCAL approach to minimize the following objectives:
\begin{equation}\label{cyclic-loss}
\begin{split}
\mathcal{L} &= \mathcal{L}_f(T, \hat{T}) + \mathcal{L}_g(Q, \hat{Q}) 
\end{split}
\end{equation}

\subsection{Multimodal Late Fusion with CCAL} \label{sec:multimodal-ccal}
We also benchmark a multimodal late fusion-based technique combined with the Cycle-Consistent Answer Localization approach as discussed in Section \ref{sec:ccal}. Our aim is to obtain visual and language modality at each word of the word sequence $W$. Towards this, in the late fusion multimodal approach, we extracted the frame (one frame per second) features corresponding to the video $V$ utilizing a 3D ConvNet (I3D) model, which was pre-trained on the Kinetics dataset \cite{carreira2017quo}. Thereafter, for each word $w \in W=\{w_1^1, \ldots w_1^{|s_1|}, \ldots, w_{n}^{1} \ldots w_{n}^{|s_n|}\}$, we obtained frame representation $F \in \mathcal{R}^{|W|\times d}$ by aligning the frame lies in the corresponding subtitle time stamp. We consider $F$ as the input image of dimension $|W|\times d$ and pass this to the vision encoder to encode the frame representation and obtain the vision representation as follows: $\mathcal{V} \leftarrow p(F; \psi)$, where $\mathcal{V} \in \mathcal{R}^{d_v}$ and $p$ is the vision encoder of parameter $\psi$. Similar to the reading comprehension model discussed in Section \ref{sec:reading-comp-model}, we obtained the hidden state representation $h_i \in \mathcal{R}^{d_l}$ and concatenated it with the vision representation $\mathcal{V}$ and obtained the multimodal representation $x_i \in \mathcal{R}^{d_l+d_v}$. Finally, we apply a feed-forward network with $relu$ activation to project the $x_i$ into language encoder dimension $d_l$ to use the pre-trained language model further as used in the reading comprehension approached discussed in Section \ref{sec:reading-comp-model}.  The multimodal late fusion with the CCAL model is trained by following the objectives listed in Eq. \ref{cyclic-loss}. In our experiments, as vision encoder $p$, we utilized VIT-Base \cite{dosovitskiy2020image}, VITMAE\cite{he2022masked}, VideoMAE-Base \cite{tong2022videomae}, VAN-Base \cite{he2022masked}, and ConvNext-Base \cite{liu2022convnet} vision-based models.

\section{Experimental Setup}

\subsection{Datasets and Implementations}
We evaluated the performance of the system on MedVidQA\cite{gupta2023dataset} dataset, which focuses on instructional visual answer localization task. There are $2,710$, $145$, and $155$ questions with visual answers annotated across the training, validation, and test sets, respectively. We also analyze the role of the created HealthVidQA-CRF dataset in improving the performance of the medical visual answer localization task. Additionally, we evaluated the best-performing approaches on the 10\% of the HealthVidQA-CRF dataset and called it the HealthVidQA-CRF test dataset. The implementation details are provided in the Appendix \ref{sec:implementations}. 

\subsection{System Evaluation}
Following the previous work, we use "R@1 IoU = $\mu$" and "mIoU" for the evaluation of visual answer localization. For each test question, we measure the Intersection over Union (IoU) between the predicted and ground truth timestamps. "R@1, IoU@ = $\mu$" means the percentage of text queries with an IoU larger than $\mu$. "mIoU" refers to the average IoU for all test questions. 



\begin{figure}[]
\begin{minipage}{.55\textwidth}
  \centering
  \resizebox{\textwidth}{!}{%
\begin{tabular}{lcccc}
\hline
\multicolumn{1}{l}{\multirow{1}{*}{\textbf{Models}}}
 & \multicolumn{1}{c}{\textbf{IoU = 0.3}} & \multicolumn{1}{c}{\textbf{IoU = 0.5}} & \multicolumn{1}{c}{\textbf{IoU = 0.7}} & \textbf{mIoU} \\ \hline \hline

VSLBase \cite{gupta2023dataset} & 25.16 & 8.38 & 4.51 & 19.3\\ 
VSLQGH \cite{gupta2023dataset} & \multicolumn{1}{c}{25.81} & \multicolumn{1}{c}{14.2} & \multicolumn{1}{c}{6.45} & 20.12 \\ 
CCGS \cite{li2022learning} & \multicolumn{1}{c}{67.1} & \multicolumn{1}{c}{50.32} & \multicolumn{1}{c}{27.74} & 47.11  \\ \hline
RC \cite{beltagy2020longformer} & \multicolumn{1}{c}{61.44} & \multicolumn{1}{c}{47.06} & \multicolumn{1}{c}{29.41} & 45.02 \\ 
CCAL (T5-QG) & \multicolumn{1}{c}{67.32} & \multicolumn{1}{c}{49.67} & \multicolumn{1}{c}{35.29} & 50.58 \\ 
CCAL & \multicolumn{1}{c}{\textbf{71.90}} & \multicolumn{1}{c}{\textbf{54.9}} & \multicolumn{1}{c}{\textbf{35.29}} & \textbf{52.92}\\ \hline
CCAL+ViT & \multicolumn{1}{c}{69.28} & \multicolumn{1}{c}{50.33} & \multicolumn{1}{c}{31.37} & 50.24 \\ 
CCAL+VideoMAE & \multicolumn{1}{c}{66.66} & \multicolumn{1}{c}{49.02} & \multicolumn{1}{c}{30.07} & 48 \\ 
CCAL+ViTMAE & \multicolumn{1}{c}{66.66} & \multicolumn{1}{c}{52.29} & \multicolumn{1}{c}{33.33} & 52.2 \\ 
CCAL+VAN & \multicolumn{1}{c}{67.97} & \multicolumn{1}{c}{47.71} & \multicolumn{1}{c}{28.1} & 48.17 \\ 
CCAL+ConvNeXt & \multicolumn{1}{c}{67.97} & \multicolumn{1}{c}{52.94} & \multicolumn{1}{c}{32.03} & 50.12 \\ 
 \hline \hline
\end{tabular}%
}
\captionof{table}{Performance compression of the multiple monomodal and multimodal approaches on MedVidQA test dataset. RC referees to the reading comprehension model. CCAL (T5-QG) denotes the CCAL approach with T5 as the question generator.}
\label{tab:medvidqa-main-results}

\end{minipage}
\centering
\quad
\begin{minipage}{.4\textwidth}
  \centering
\resizebox{\textwidth}{!}{%
\begin{tabular}{llcccc}
\hline
 & \textbf{Models} & \multicolumn{1}{c}{\textbf{IoU = 0.3}} & \multicolumn{1}{c}{\textbf{IoU = 0.5}} & \multicolumn{1}{c}{\textbf{IoU = 0.7}} & \multicolumn{1}{c}{\textbf{mIoU}} \\ \hline \hline
1 &RC \cite{beltagy2020longformer} & \multicolumn{1}{c}{61.44} & \multicolumn{1}{c}{47.06} & \multicolumn{1}{c}{29.41} & 45.02 \\ 
2 & \quad + 10\% HeathVidQA-CRF & 66.01 & 52.29 & 38.56 & 51.81 \\ 
3 & \quad + 20\% HeathVidQA-CRF  & 67.32 & \textbf{54.25} & \textbf{35.95} & \textbf{51.84} \\ 
4 & \quad 
+ 50\% HeathVidQA-CRF & \textbf{68.63} & {52.94} & 33.99 & 51.6 \\ \hline 
\\
5 & CCAL & \multicolumn{1}{c}{\textbf{71.90}} & \multicolumn{1}{c}{\textbf{54.9}} & \multicolumn{1}{c}{35.29} & \textbf{52.92}\\
6 & \quad + 10\% HeathVidQA-CRF & 64.05 & 47.71 & 34 & 48.03 \\ 
7 & \quad + 20\% HeathVidQA-CRF & 69.28 & 52.94 & 35.95 & 51.72 \\ 
8 & \quad + 50\% HeathVidQA-CRF & 66.01 & 52.94 & \textbf{36.60} & 51.1 \\ \hline \hline
\end{tabular}%
}
\captionof{table}{Effect of the portion of the HealthVidQA-CRF on the performance of the MedVidQA test dataset.}
  \label{tab:medvidqa-results-with-healthvidqa-portion}
\end{minipage}
\end{figure}

\section{Results and Analysis}
We evaluated the multiple monomodal and multimodal approaches discussed in Section \ref{sec:approaches} on the MedVidQA and HealthVidQA test datasets. The detailed results are presented in the Table \ref{tab:medvidqa-main-results}. We first evaluated the performance of the reading comprehension (RC) model (\textit{cf.} Section \ref{sec:reading-comp-model}) on the MedVidQA test dataset and reported the IoU=0.7 as $29.41$, which is the strictest measure to evaluate the predicted visual answer segment. The RC model achieved the mIoU value of $45.02$. Our proposed CCAL model achieved the IoU=0.7 and mIoU values as $35.29$ and $52.92$, respectively. The CCAL model reported the $5.88$ and $7.9$ absolute performance gain compared to the RC model. It is to be noted that the RC model is trained only by minimizing the $\mathcal{L}_f(T, \hat{T})$ loss, while the CCAL model is trained by minimizing the  $\mathcal{L}_f(T, \hat{T})$ as well as $\mathcal{L}_g(Q, \hat{Q})$. The performance improvement with the CCAL model over the RC model signifies the importance of the question generation component, which enforce the RC model to predict such answer for which the generated question and ground truth questions are semantically and syntactically similar. In our proposed CCAL approach, we used BART as a question generator. To assess its effectiveness, we replaced BART with the T5 model, called it as CCAL (T5-QG) model, and reported the performance in Table \ref{tab:medvidqa-main-results}. We observed absolute performance decrement of $4.58$, $5.23$, and $2.34$ in terms of IoU=0.3, IoU=0.5, and mIoU, respectively. 

We have also explored the role of visual features in detecting visual answer segments to medical instructional questions. Toward this, we experiment with multiple visual encoders along with the CCAL model, which makes the resultant architecture multimodal. Amongst all the multimodal approaches, CCAL+ViT obtained the maximum score of 69.28 for the IoU=0.3 metric. Similarly, the models CCAL+ViTMAE achieved the maximum score of  $33.33$, and $52.2$ for the IoU=0.7 and mIoU metrics. For the metric IoU=0.5, the model CCAL+ConvNeXt achieved the maximum score of $52.94$. We also observed that none of the multimodal approaches could outperform the CCAL model. CCAL+ViTMAE was closest to CCAL in terms of IoU=0.7 and mIoU metrics.
\begin{figure}[]
\begin{minipage}{.55\textwidth}
  \centering
  \resizebox{\textwidth}{!}{%
\begin{tabular}{llcccc}
\hline
 & \textbf{Models} & \textbf{IoU = 0.3} & \textbf{IoU = 0.5} & \textbf{IoU = 0.7} & \textbf{mIoU} \\ \hline \hline
1 & CCAL+VAN & \multicolumn{1}{c}{\textbf{67.97}} & \multicolumn{1}{c}{47.71} & \multicolumn{1}{c}{28.1} & 48.17 \\ 
2 & \quad + 10\% HeathVidQA-CRF & {66.67} & \textbf{52.29} & \textbf{38.56} & \textbf{52.25} \\ \hline
\\
3 & CCAL+ConvNeXt & \multicolumn{1}{c}{67.97} & \multicolumn{1}{c}{52.94} & \multicolumn{1}{c}{32.03} & 50.12 \\
4 &  \quad + 10\% HeathVidQA-CRF & 62.09 & 47.06 & 30.72 & 47.54 \\ 
5 & \quad + 20\% HeathVidQA-CRF& \textbf{69.28} & 54.25 & 30.72 & 50.60 \\ 
6 &  \quad + 50\% HeathVidQA-CRF& 66.01 & 49.67 & 29.41 & 48.56 \\
7 &  \quad + 100\% HeathVidQA-CRF & 67.32 & \textbf{54.90} & \textbf{37.91} & \textbf{51.77} \\ \hline \hline
\end{tabular}%

}
\captionof{table}{Effect of the portion of the HealthVidQA-CRF on the performance of the multimodal approaches on the MedVidQA test dataset.}
\label{tab:medvidqa-multimodal-results-with-healthvidqa-portion}

\end{minipage}
\centering
\quad
\begin{minipage}{.4\textwidth}
  \centering
\resizebox{\textwidth}{!}{%
\begin{tabular}{llcccc}
\hline
 & \textbf{Models} & \textbf{IoU = 0.3} & \textbf{IoU = 0.5} & \textbf{IoU = 0.7} & \textbf{mIoU} \\ \hline \hline
1 &RC \cite{beltagy2020longformer} & 
\multicolumn{1}{c}{{51.11}} & \multicolumn{1}{c}{{32.51}} & \multicolumn{1}{c}{17.30} & {36.39}\\
2 & \quad + 10\% HeathVidQA-CRF  & 63.01 & 41.32 & 23.39 & 44.76 \\ 
2 & \quad + 20\% HeathVidQA-CRF  & {66.41} &{44.68} & {25.72} & {46.96} \\ 
3 & \quad + 50\% HeathVidQA-CRF  & 68.52 & 46.29 & 26.74 & 48.01 \\ 
4 & \quad + 100\% HeathVidQA-CRF & \textbf{70.04} & \textbf{49.82} & \textbf{30.01} & \textbf{50.35} \\ \hline \\
5 & CCAL & \multicolumn{1}{c}{53.67} & \multicolumn{1}{c}{32.82} & \multicolumn{1}{c}{17.35} & 37.13 \\
6 & \quad + 10\% HeathVidQA-CRF & 62.03 & 39.53 & 22.41 & 43.48 \\ 
7 & \quad + 20\% HeathVidQA-CRF & 66.73 & 45.84 & 26.61 & 47.22 \\ 
8 & \quad + 50\% HeathVidQA-CRF & \textbf{68.16} & \textbf{46.33} & \textbf{27.78} & \textbf{48.12} \\ \hline \\

9 & CCAL+ConvNeXt & \multicolumn{1}{c}{52.45} & \multicolumn{1}{c}{30.84} & \multicolumn{1}{c}{16.21} & 35.19 \\
10  & \quad + 10\% HeathVidQA-CRF & 56.57 &	36.18	&20.04	&40.48\\  
 11 & \quad + 50\% HeathVidQA-CRF & 69.14 &	47.99&	29.65&	49.1\\ 
12 & \quad + 100\% HeathVidQA-CRF& \textbf{72.05} &	\textbf{50.4} &	\textbf{30.68}	&\textbf{51.07}\\ \hline \hline
\end{tabular}%
}

\captionof{table}{Effect of the portion of the HealthVidQA-CRF on the performance of the HealthVidQA-CRF test dataset.}
  \label{tab:healthvidqa-results-with-healthvidqa-portion}
\end{minipage}
\end{figure}

\subsection{Effect of HealthVidQA-CRF Dataset}
We assess the effect of the created HealthVidQA-CRF dataset on the models trained and evaluated (\textit{cf.} Table \ref{tab:medvidqa-results-with-healthvidqa-portion}) on the MedVidQA test dataset. We chose the RC and CCAL models, which are the best-performing models on the MedVidQA test dataset. We begin by adding $10\%$ of the created HealthVidQA-CRF dataset into the training set of MedVidQA and trained the RC models. With the $10\%$ addition of the HealthVidQA-CRF dataset, we observe the absolute improvements of the $4.57$, $5.23$, $9.15$, and $6.79$ in terms of IoU=0.3, IoU=0.5,  IoU=0.7 and mIoU metrics, respectively. The significant improvements signify that the created dataset is capable of providing additional informative samples, which is required to train an efficient visual answer localization system. The introduction of the additional datasets improves the performance further on all the evaluation metrics. We obtained the maximum value ($68.63$) of IoU=0.3 with $50\%$ of the HealthVidQA-CRF dataset. For the remaining metrics, $20\%$ of the HealthVidQA-CRF dataset obtained the maximum scores of $54.25$ IoU=0.5, $35.95$ IoU=0.7, and $51.84$ mIoU. We followed the same for the CCAL model, too, and reported the results in Table \ref{tab:medvidqa-results-with-healthvidqa-portion}. We trained the CCAL model by adding a portion of the HealthVidQA-CRF dataset into the MedVidQA training set. We observe an absolute improvement of $1.31$ in terms of the strictest metric IoU=0.7 with a 50\% addition of the HealthVidQA-CRF dataset. 
\subsection{Effect of Visual Features with HealthVidQA-CRF Dataset}
In another analysis, we aim to assess the effect of the visual features while adding a portion of the HealthVidQA-CRF dataset to train the visual answer localization system. Table \ref{tab:medvidqa-main-results}, shows that multimodal approaches could not outperform the best-performing monomodal CCAL approach. We observed from Table \ref{tab:medvidqa-main-results} that CCAL+VAN obtained the lowest scores in terms of multiple evaluation metrics. We wanted to analyze the effect of the HealthVidQA-CRF dataset on this model. Toward this, we train the CCAL+VAN model with the 10\% of the HealthVidQA-CRF dataset. The trained model achieved an absolute improvement of $4.58$, $10.46$, and $4.08$ in terms of IoU=0.5, IoU=0.7, and mIoU evaluation metric, respectively. In a similar way, we repeated the experiments with the CCAL+ConvNeXt model and obtained an absolute improvement of $1.31$ IoU=0.3 with 20\% of the HealthVidQA-CRF dataset. With the full HealthVidQA-CRF dataset, we achieved scores of $54.90$ IoU=0.5, $37.91$  IoU=0.7 mIoU, which shows an improvement of $2.61$  IoU=0.5, $5.88$  IoU=0.7, and $1.65$ mIoU over the CCAL+ConvNeXt model. Furthermore, with this additional dataset, the CCAL+ConvNeXt model shows an absolute improvement of $2.62$ over the best-performing CCAL model in terms of the strictest metric IoU=0.7.  These significant improvements signify that multimodal approaches need additional datasets to perform better on the task of visual answer localization. 
\subsection{Performance on HealthVidQA-CRF Dataset}
We extend our experiments by evaluating the best-performing monomodal and multimodal on the test set of the HealthVidQA-CRF dataset. We performed these experiments in data incremental setup, where we first utilized the MedVidQA training set to train the model and validated its performance on the HealthVidQA-CRF test dataset. Thereafter, we added the HealthVidQA dataset in an incremental manner and analyzed its impact on the model's performance towards the HealthVidQA-CRF test dataset. We evaluated the performance of the RC model on the HealthVidQA test dataset and reported the results in Table \ref{tab:healthvidqa-results-with-healthvidqa-portion}. Thereafter, we trained the model with an additional 10\%, 20\%, 50\%, and 100\% of the HealthVidQA dataset along with the MedVidQA training set and obtained the results. We analyzed that the model achieved maximum scores of $70.04$ IoU=0.3, $49.82$ IoU=0.5, and $30.01$ IoU=0.7, and $50.35$ mIoU with 100\% of the HealthVidQA dataset. We observed the $18.93$, $17.31$, $10.71$, and $13.96$ improvements in terms of IoU=0.3, IoU=0.5, IoU=0.7, and mIou, respectively, over the RC model trained with only MedVidQA training dataset. We continued the same experimental setup for the CCAL approach as well and recorded the $14.49$, $13.51$, $10.43$, and $10.99$ improvements with 50\% of the HealthVidQA dataset, in terms of IoU=0.3, IoU=0.5, IoU=0.7, and mIou, respectively. Table \ref{tab:medvidqa-multimodal-results-with-healthvidqa-portion} shows the effectiveness of using an additional HealthVidQA-CRF dataset to train the CCAL+ConvNeXt multimodal approach. In this direction, we followed the same experimental setup and evaluated the performance CCAL+ConvNext model on the HealthVidQA test dataset. The experimental results show that with the additional HealthVidQA dataset, the CCAL+ConvNeXt model outperformed the RC and CCAL approaches. On the evaluation metric, IoU=0.5 (IoU=0.7), we recorded absolute improvement of $19.56$ ($14.47$) over the best-performing CCAL monomodal approach.

\section{Conclusion}
In this work, we presented a pipeline to automatically create medical visual question-answering datasets focusing on health-related questions and their visual answers in the videos. With the proposed pipeline, we build two large-scale medical visual question-answering datasets, HealthVidQA-CRF and HealthVidQA-Prompt. We performed in-depth human evaluations on the created datasets, and the evaluation shows the former dataset is better aligned with the human annotations. We also proposed monomodal and multimodal approaches for medical video question-answering task that achieved state-of-the-art performances and set competitive baselines for future research. The detailed experiments and analysis show that the created datasets help in improving the performance of the MedVidQA system. We believe that the created datasets can be used to provide the solution by pre-training/fine-tuning language-vision models for medical visual answer localization task.

\section*{Ethics Statement}
{\bf Dataset generation and models}: The datasets presented in this work leverage the existing publicly available HowTo100M dataset. The subsets generated through leveraging the HowTo100M dataset may inherit gender, age, geographical, and cultural biases from the public instructional videos available online. General ethical considerations that apply
to visual recognition models may apply to our models. \\
{\bf Limitations}: The health topics covered in our datasets are limited to those that are available in the HowTo100M dataset videos, which may leave some health topics underexplored.  

\section*{Acknowledgments}
This work was supported by the intramural research program at the U.S. National Library of Medicine, National
Institutes of Health, and utilized the computational resources of the NIH HPC Biowulf cluster (\url{http://hpc.nih.gov}).
The content is solely the responsibility of the authors and does not necessarily represent the official views of the National Institutes of Health. We would like to thank Debora Whitman, Denise Hunt, Melanie Huston, Dorothy Trinh, Preeti Kochar, Cathy Smith, Luke Zic, and Arun Crispino for their annotations and contributions to this work.
\bibliographystyle{plainnat}
\bibliography{references}  
\appendix

\section{Implementation Details} \label{sec:implementations}
\subsection{Dataset Creation}
\subsubsection{Selecting Medical Instructional Videos}
We use HuggingFace \cite{wolf2019huggingface} implementations of the BigBird$_{\text{BASE}}$ model and train it with the batch size of $4$ with a maximum token length of $1024$. The model parameters are updated using the Adam \cite{kingma2015adam} optimization algorithm with the learning rate of $5e-5$. We obtained the value of the optimal hyperparameters based on the MedVidCL validation dataset performance in terms of the F1-score.
\subsubsection{Detecting Visual Answer Segments}
 We utilized the pre-trained models from the HuggingFace model hub. We use the base version of all the pre-trained language models to detect visual segments. The position embedding is set to the dimension of the hidden state of the respective pre-trained language models. We set the maximum token length of each segment to $128$. The models are trained with a batch size of $4$ with one layer of transformer encoder. The model parameters are updated using the AdamW \cite{loshchilov2018decoupled} optimization algorithm with the learning rate of $4e-5$ and weight decay of $1e-4$.

 \subsubsection{Generating Medical Instructional Questions}
For question generation, we use the large version of pre-trained T5, BART, and PEGASUS models. The transformer-based encoder-decode model was trained with one layer of encoder and decoder, each with a hidden state size of $128$. We use the official repository\footnote{\url{https://github.com/microsoft/UniVL}} of UniVL with the default hyper-parameters to fine-tune for the question generation task.
The pre-trained language models are trained with a batch size of $2$ with a source sequence length of $256$ and question generation target sequence length of $20$. We use the beam search to generate the question with beam size=5. The model parameters are updated using the AdamW \cite{loshchilov2018decoupled} optimization algorithm with the learning rate of $4e-5$ and weight decay of $1e-4$.
We use the evaluation metrics implemented in the dataset library\footnote{\url{https://github.com/huggingface/datasets}}.
\subsection{Approaches}
We utilized the base version, pre-trained language, and vision models (T5, BART, Longformer, ViT, VideoMAE, ViTMAE, VAN, and ConvNeXt) from HuggingFace \cite{wolf2019huggingface} to perform the experiments.
For RC and CCAL approaches, we set the maximum source sequence length to 1024, except for the Longformer model. The Longformer model was set to 4096.
 For visual features, we select one frame from each second of the video uniformly and extract RGB visual features with the 3D ConvNet that was pre-trained on the Kinetics dataset \cite{carreira2017quo}.
 Each pre-trained Transformer model was trained with the AdamW optimizer with the learning rate=$5e-5$ for ten epochs, along with early stopping with the patience of 3 epochs and a batch size of 2. 
\end{document}